\documentclass[runningheads]{llncs}
\usepackage[T1]{fontenc}
\usepackage{graphicx}
\usepackage{booktabs}
\usepackage{cite}
\begin{document}
\title{AutoMode-ASR: Learning to Select ASR Systems for Better Quality and Cost}
\titlerunning{AutoMode ASR}
%
\author{Ahmet G\"und\"uz\and
Yunsu Kim\and
Kamer Ali Yuksel\and Mohamed Al-Badrashiny\and Thiago Castro Ferreira\and Hassan Sawaf}
\authorrunning{A. G\"und\"uz et al.}
%
\institute{aiXplain Inc., San Jose, CA, USA \\
\email{\{ahmet,yunsu.kim,kamer,mohamed,thiago,hassan\}@aixplain.com}\\
\url{https://aixplain.com/}}
\maketitle              
\begin{abstract}
We present AutoMode-ASR, a novel framework that effectively integrates multiple ASR systems to enhance the overall transcription quality while optimizing cost.
The idea is to train a decision model to select the optimal ASR system for each segment based solely on the audio input before running the systems.
We achieve this by ensembling binary classifiers determining the preference between two systems.
These classifiers are equipped with various features, such as audio embeddings, quality estimation, and signal properties.
Additionally, we demonstrate how using a quality estimator can further improve performance with minimal cost increase.
Experimental results show a relative reduction in WER of 16.2\%, a cost saving of 65\%, and a speed improvement of 75\%, compared to using a single-best model for all segments.
Our framework is compatible with commercial and open-source black-box ASR systems as it does not require changes in model codes.

\keywords{Automatic speech recognition \and Quality estimation \and Cost optimization.}
\end{abstract}
\section{Introduction}
Automatic speech recognition (ASR) has evolved remarkably due to advances in deep learning \cite{graves2013speech,chan2016listen,gulati2020conformer}.
Consequently, numerous high-quality ASR models have been released \cite{zhang2023google,pratap2023scaling,radford2023robust}, with some claiming to achieve human parity.

However, users frequently encounter challenges in selecting the most suitable ASR model for their speech data; the performance of various models can differ on the same segment, and their rankings may vary depending on the input conditions, such as accents, dialects, background noises, and speaking styles \cite{benzeghiba2007automatic}.
For instance, certain models are optimized for studio recordings, while others are more robust to non-speech noise.
It is important to note that audio conditions can vary across different segments, even within the same corpus and application.

This variability poses a significant challenge in intelligently integrating multiple ASR systems for a specific purpose.
Traditionally, this has been addressed through system combination \cite{fiscus1997post,mangu2000finding}, which constructs a confusion network from multiple hypotheses and finds the best path to derive the final transcription.
Ensemble learning methods introduce diversity among the systems to expand the combination space \cite{schwenk1999using,siohan2005constructing}.
Departing from the confusion network, \cite{gitman23confidence} propose leveraging confidence scores from ASR systems to select the optimal hypothesis.
While effective in reducing the Word Error Rate (WER), these approaches share a common limitation: they necessitate hypotheses from all candidate systems.
Meeting this requirement is often impractical nowadays due to the large model size of high-performing modern ASR systems; commercial systems are expensive, and open-source models incur substantial costs. 

This paper introduces AutoMode-ASR, a novel framework that predicts the most suitable ASR system for a given audio segment—defined as a contiguous chunk of speech, such as a sentence or phrase—without running the inference of candidate systems.
The prediction is based on features extracted from the audio input, and its transcription is performed only with the predicted system afterward.
The distinct separation between system selection and inference eliminates the requirement of modifying the decoding process, enabling a flexible combination of commercial and open-source models.

Our experimental results show that AutoMode-ASR improves transcription performance by up to -16.2\% relative in WER.
Notably, compared to other multi-system approaches, it does not increase operational costs; rather, it reduces costs by opting for a lighter system in cases of comparable performance, achieving a price reduction of 65\%. 
Our contributions are:
\begin{itemize}\itemsep0em
    \item We present a new combination scheme for any ASR models at the segment level to optimize quality and cost.
    \item We analyze feature types to discern their relevance in accurately predicting the performance of ASR systems.
    \item We propose a robust classification module that facilitates the incremental integration of ASR systems.
    \item We demonstrate an effective approach to incorporate quality estimation for further optimizing performance.
\end{itemize}

\begin{figure}[ht]
\centering
\includegraphics[width=0.7\textwidth]{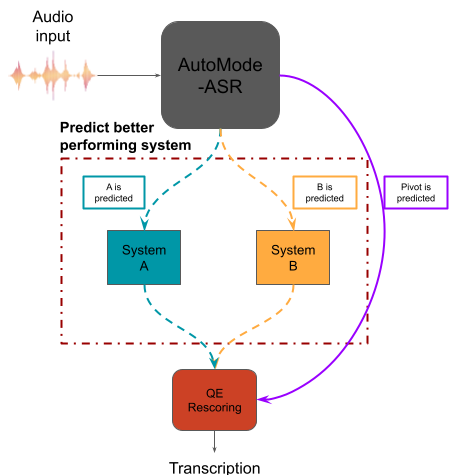}
\caption{The Diagram of the AutoMode-ASR Workflow}
\label{fig:workflow}
\end{figure}

\section{Methodology}
AutoMode-ASR aims to predict the optimal ASR system for a given audio input, a task framed as multi-class classification using system IDs as class labels.
Training a classifier for this involves preparing data by conducting ASR inference for each candidate system on every audio segment, entailing significant costs. 
If thus only a limited amount of training data can be prepared, there is a high risk of insufficient cases for certain systems as top performers.
This class imbalance adversely impacts classification accuracy \cite{ou2007multi,wang2012multiclass}, although it can be mitigated by data sampling \cite{chawla2002smote,abdi2015combat} or boosting methods \cite{yijing2016adapted,tanha2020boosting}.

In this work, we approach the problem as learning to rank \cite{liu2009learning}, leveraging the inherent rank information within our training data.
Following the ASR inferences on a training segment, we acquire not only the ID of the top-performing system but also the rank across all candidates, sorted by WER.
In contrast to classification, which is trained solely on the top system ID, ranking gives training instances for every system from each segment with relative performance.
This maximizes the utilization of information within the training data and enhances stability.

\subsection{Method Overview}

Rather than directly predicting a ranking over all systems, we decompose the problem into multiple binary classification problems (Section \ref{sec:traning}), each comparing a specific pair of systems \cite{hullermeier2008label}.
As mentioned, every training segment has ranking information between candidate systems.
To train a binary classifier, we simply relabel each segment with the winning system between the two systems in question.
Each binary classifier is then trained with all available segments, which is critical in our setup where data preparation is difficult.

Following the one-vs-one approach, learning to rank pairwisely ideally requires considering every pair of systems: $C(C-1)/2$ pairs with $C$ as the number of systems.
However, this proves impractical in numerous scenarios, leading to the development of various pair sampling methods \cite{wauthier2013efficient,shah2018simple}.
In our method, we designate a \emph{pivot} system from all candidates, playing the role of a comparative baseline against every other system.
Each binary classifier compares the pivot against another candidate, resulting in $C-1$ pairs, referred to as \emph{one-vs-pivot}.
Among multiple systems, we strategically choose the cost-effective system as the pivot.
This decision introduces an implicit bias towards minimizing costs, aligning with our goal.

The decisions from individual binary classifiers are later merged into the final decision using a simple heuristic (Section \ref{sec:ensemble}).
This two-pass strategy is advantageous when a new system is incrementally added to the comparison.
In such cases, training another binary classifier between the pivot system and the new system suffices, which is significantly more efficient than retraining a multi-class classifier involving all systems.
Figure \ref{fig:workflow} shows the entire workflow of the proposed method.

As our binary classifier, we employ the Gradient Boosting Machine (GBM) \cite{friedman2001greedy} due to its flexibility in feature integration and interpretability in analyzing feature importance.
GBMs have demonstrated superiority over deep neural networks across various classification and ranking problems, excelling in terms of both accuracy and efficiency \cite{ebrahimi2019comprehensive,schmitt2022deep}.
The GBM algorithm incrementally incorporates weak learners by fitting each new learner to the residuals of the preceding ensemble.
Specifically, we utilize the eXtreme Gradient Boosting (XGBoost) package \cite{chen2016xgboost}, where each weak learner is a regression tree.

\subsection{Features}
\label{sec:binary:features}

An essential research question in our work is identifying relevant aspects of audio input for predicting ASR performance.
To study this question, we integrate diverse features in the binary classifiers designed to encapsulate audio files' acoustic, linguistic, and quality-related dimensions:

\begin{itemize}
  \item \textbf{Self-supervised audio embedding}:
  Representations learned from speech audio via contrastive learning on the masked latent space prove beneficial for many downstream speech tasks \cite{baevski2020wav2vec}.
  We adopt the cross-lingual version of it, trained on 53 languages (Wav2Vec2-XLSR-53) \cite{conneau2020unsupervised}.
  We extract the last encoder states and average them over the time dimension to produce a consolidated vector of 1024 dimensions.
  \item \textbf{Input language}:
  AutoMode-ASR operates without assuming the audio input language; it can accommodate any supported languages across all systems.
  Recognizing that classifier decisions may differ based on language, we include the language of each speech segment as a categorical feature.
  This empowers the classifier to potentially select the model that performs better in that particular language.
\end{itemize}

In addition to extracting features directly from the audio file, we utilize a lightweight ASR model to capture valuable features from the inference process and its temporary transcription.
Note that this ASR model differs from the systems compared within AutoMode-ASR.
We opt for a compact and swift ASR model for feature extraction, thereby minimizing any significant increase in processing time.
After the ASR inference, we obtain the following features:

\begin{itemize}
  \item\textbf{ASR embedding}:
  We extract the output states of the encoder from the ASR model and compute their average over time.
  Our hypothesis posits that the representations learned during transcription encapsulate precise information relevant to estimating transcription performance.
  \item \textbf{ASR confidence score}:
  ASR inference provides log probabilities for each output token, called confidence scores.
  We include the mean, standard deviation, and five-number summary of the probability values as features.
  These scores offer preliminary insights into the transcription's difficulty level; some systems may excel in deciphering ambiguous phonetics in challenging segments, while others may perform better in transcribing easier segments. 
  \item \textbf{Quality estimation score}:
  Once we obtain a transcription from the feature-extracting ASR model, we can assess its quality even without a reference using quality estimation metrics; it serves as another indicator of speech recognizability.
  For this purpose, we utilize \emph{NoRefER} \cite{yuksel23_icassp,yuksel23_interspeech,javadi2024wordlevel}, which computes a score for the transcription and exhibits a high correlation with WER.
  It is worth noting that the NoRefER score is calculated solely based on the transcription itself, without considering the audio, thus providing a distinct dimension of information compared to other features.
  \item\textbf{Quality estimation embedding}:
  NoRefER itself is a neural network, from which we can extract representations from its intermediate layers. Specifically, we extract the last hidden state before the final linear layer of the NoRefER network, resulting in an embedding of 384 dimensions.
\end{itemize}

\subsection{Training}
\label{sec:traning}
For training a binary classifier between the pivot and any ASR system, we first assign labels to each segment based on the system with the lower WER between the two.
In cases where WER values are identical, we label them with the more cost-effective system.
We observed numerous instances with identical WER values, particularly for short segments, resulting in substantial cost savings overall (Section \ref{sec:exp:main}).

To enhance training and increase the WER gain, we prioritize samples with carefully crafted sample weights, calculated as the product of the following factors:
\begin{itemize}
    \item \textbf{Normalized WER difference}: Calculate the absolute WER difference between two systems and divide each value by the range of values across the entire training set.
    Segments with a larger difference in WER are given more weight in loss calculation as correctly classifying these segments is expected to yield greater gains.
    \item \textbf{Inverse label frequency}: To counteract bias toward a single system, we assigned higher weights to the minority label.
\end{itemize}

Classifier training minimizes the binary logistic loss along with its first and second-order gradients at each step of adding a weak regression tree \cite{chen2016xgboost}.
The hyperparameters of the trees, such as the number of leaves, number of features, or learning rate, are selected using cost-frugal hyperparameter optimization \cite{wu2021frugal}, which samples a tree learner based on the estimated cost for improvement.
Each hyperparameter setting is evaluated using cross-validation with WER reduction, i.e., WER decrease by AutoMode-ASR selections versus selecting a single system.

\subsection{Multi-Class Ensemble}
\label{sec:ensemble}

For comparing all systems, we aggregate predictions from individual binary classifiers and choose the most confident decision.
Every binary classifier provides a prediction and the probability between its two systems, which exceeds 0.5.
We select the system predicted with the highest probability among multiple binary classifiers. Note that each binary classifier compares a system with the pivot system.
If a system surpasses the pivot, it is selected; if none of the systems outperform the pivot, we default to the cost-effective pivot system.


For more elaborate decision-making and further cost savings, AutoMode-ASR offers an option to rescore comparisons when a more expensive system is chosen, i.e., the pivot is not selected.
We obtain transcriptions from the systems and compute their quality estimation scores using NoRefER, ultimately selecting the system with the highest score.
Since the comparison is based on system outputs and the WER-correlated NoRefER, we anticipate that this yields predictions more aligned with WER, while also providing another opportunity for the low-cost pivot to be selected. In contrast to the features involved in the initial decision (Section \ref{sec:binary:features}), this process requires running the ASR systems and comparing their transcriptions. The only extra cost is due to the NoRefER inference.

\section{Experiments}

For our experiments, we curated training and test data by selecting diverse audio samples from Common Voice \cite{ardila2020common} and LibriSpeech \cite{panayotov2015librispeech}.
Combining these two sources, our dataset encompasses various speaking styles and recording acoustics.
We included the English, French, Spanish, German, and Russian subsets from Common Voice to thoroughly evaluate AutoMode-ASR's adaptability and effectiveness across different languages.
Each selected segment was then inputted into all systems under comparison to obtain the WER and the performance ranking.
The statistics of the prepared data are in Table \ref{tab:data}.

\begin{table}[!ht]
    \centering
    \caption{The number of audio segments where each ASR system (System A, System B, System C, Whisper) achieved the best performance in terms of Word Error Rate (WER) in training, validation, and testing subset.}
    \label{tab:data}
    \begin{tabular}{cccc}
    \toprule
    Top-Rank System & \texttt{train} & \texttt{valid} & \texttt{test} \\
    \midrule
    System A & \enspace1,182 & \enspace\,149 & \enspace\,149 \\
    System B & \enspace1,322 & \enspace\,189 & \enspace\,174 \\
    System C & \enspace5,431 & \enspace\,735 & \enspace\,773\\
    Whisper (pivot) & 14,817 & 2,147 & 2,107\\
    \midrule
    Total & 22,752 & 3,220 & 3,203 \\
    \bottomrule
    \end{tabular}
\end{table}

We evaluated AutoMode-ASR using three commercial ASR systems (called System A/B/C\footnote{The disclosure of the system providers is pending approval under legal review. Their names will be disclosed accordingly after the review.}) alongside Whisper \cite{radford2023robust}, an open-source model developed by OpenAI.
These providers were selected due to their prominence in the industry and diverse speech recognition approaches, ensuring a comprehensive and practical evaluation.
Whisper, specifically its ``small'' version, was chosen as the pivot among systems because of its low cost and latency. In feature extraction (Section \ref{sec:binary:features}), we employed the Whisper small model for ASR embedding and confidence score.
While this choice coincides with the pivot model, it was made solely for our convenience and does not introduce any unintended bias toward the pivot in AutoMode-ASR.

To train the classifiers, we employed Microsoft's FLAML framework \cite{wang2021flaml}.
We conducted five-fold cross-validation with a time budget of 1,000 seconds for hyperparameter optimization (HPO).
While we also involved LightGBM \cite{ke2017lightgbm} and CatBoost \cite{prokhorenkova2018catboost} machines in the HPO process, XGBoost consistently outperformed the others; thus we only present the results obtained using XGBoost.
The weighting scheme for training data sampling and the target metric for HPO are tuned according to the performance on a validation set.

System selection was assessed against the top-ranking systems using F1, weighted by inverse label frequency to address label imbalance (Table \ref{tab:data}).
Subsequently, ASR decoding was conducted with the selected system per segment, and WER was computed to evaluate AutoMode-ASR's actual improvement in transcription performance.
These results were compared against selecting one system for all segments: the pivot system (pivot only), a non-pivot system (non-pivot only), or the system with the best overall performance when used for all segments (single-best).
We removed punctuation marks and applied lowercasing before computing WER.

\subsection{Main Results}
\label{sec:exp:main}

Table \ref{tab:result:binary} shows the reduction in WER and classification performance of each binary classifier.
While the pivot system is competitive, it does not consistently outperform other commercial systems in all segments, particularly compared to System C. Even though Table \ref{tab:data} shows that the pivot outperforms System C in nearly three times as many segments, System C achieves a lower average WER than the pivot (Whisper), likely because System C excels in more challenging segments where WER reduction has a greater impact.

AutoMode-ASR's binary classifiers efficiently identify cases where alternative systems excel over the pivot, showcasing the framework's ability to optimize ASR system selection.
Sample weighting in training (Section \ref{sec:traning}) consistently proves beneficial and QE rescoring provides an additional gain.

Table \ref{tab:result:multiclass} displays the final results after ensembling all three binary classifiers.
Compared to selecting a single system for all segments, AutoMode-ASR achieves significantly lower WER, decreasing from 13.4\%  to 11.6\%, with QE rescoring further reducing it to 11.1\%.
Notably, this improvement does not increase operational costs or delays; it requires approximately 36\% of the cost and 25\% of the runtime of the single-best baseline.
It is noteworthy that QE rescoring only introduces negligible extra cost and runtime.
A small open-source model could nearly eliminate both the cost and runtime by selecting the pivot system exclusively. However, its performance is significantly inferior to that of AutoMode-ASR, as it does not benefit from strong commercial systems.
Additionally, we provide the performance metrics when using actual top-performing systems (``Oracle''), indicating potential room for improvement in future work.

These figures not only highlight the system's processing efficiency but also its cost-effectiveness compared to the baseline 'Single-best' system.

\begin{table*}[!t]
\centering
\caption{Word Error Rate (WER) reduction and classification performance (F1-score) for each \textbf{binary classifier} comparing the pivot system (Whisper) against commercial systems (A, B, C). Results are shown for different system selection strategies, including AutoMode-ASR w/wo sample weighting and quality estimation (QE) rescoring.}
\label{tab:result:binary}
\begin{tabular}{lcccccc}
\toprule
 Pivot vs. & \multicolumn{2}{c}{System A} & \multicolumn{2}{c}{System B} & \multicolumn{2}{c}{System C} \\ 
 \cmidrule(r{2pt}){1-1}\cmidrule(l{1pt}r{2pt}){2-3}\cmidrule(l{2pt}r{2pt}){4-5}\cmidrule(l{2pt}r{2pt}){6-7}
 & WER & F1 & WER & F1 & WER & F1\\
 System Selection  & [\%] & [\%] & [\%] & [\%] & [\%] & [\%] \\
\midrule
Non-pivot only                         & 21.2 & \enspace5.7  & 20.8 & \enspace4.7 & 13.4 & 11.8\\
Pivot only                          & 14.1 & 73.4 & 14.1 & 75.9 & 14.1 & 61.1\\
\midrule
AutoMode-ASR                        & 13.6 & 77.8 & 14.0 & 76.9 & 12.2 & 72.9\\
 + Sample weights                   & 13.1 & 79.0 & 13.9 & 78.1 & \textbf{11.3} & 73.1\\
 \enspace + QE rescoring            & \textbf{12.9} & \textbf{80.2} & \textbf{13.7} & \textbf{79.1} & 11.4 & \textbf{76.3}\\
\bottomrule
\end{tabular}
\end{table*}

\begin{table*}[!t]
  \caption{Multi-class ensemble results comparing Word Error Rate (WER), F1 score, cost, and runtime of different system selection strategies. The "Single-best" system represents the baseline. AutoMode-ASR and its variants with sample weighting and quality estimation (QE) rescoring achieve progressively lower WER at a reasonable decrease in cost and runtime. The "Oracle"  represents the perfect prediction scenario.}
  \centering

  \label{tab:result:multiclass}
  \begin{tabular}{lcccc}
    \toprule
     & WER & F1 & Cost & Runtime\\
    System Selection & [\%] & [\%] & [\%] & [\%] \\
    \midrule
    Single-best         & 13.4 & \enspace\enspace9.4& 100.0 & 100.0\\
    Pivot only          & 14.1 & \enspace52.2& \enspace\enspace\textbf{2.3}& \enspace\enspace\textbf{4.6}\\
    \midrule
    AutoMode-ASR    & 12.3 & \enspace62.5 & \enspace18.6 & \enspace19.3\\
    + Sample weights         & 11.6 & \enspace63.4 & \enspace36.2 & \enspace24.9\\
    \enspace + QE rescoring     & \textbf{11.1} & \enspace\textbf{65.5} & \enspace36.2 & \enspace25.1\\
    \midrule
    Oracle & \enspace6.5 & 100.0 & \enspace41.0  & \enspace37.4 \\
    \bottomrule
  \end{tabular}
\end{table*}

\subsection{Feature Ablation}

\begin{table}[!t]
  \centering
  \caption{AutoMode-ASR results with different feature groups w/o QE rescoring, and when QE rescoring is applied to the best setting (all). ``{\normalfont Audio}'' stands for self-supervised audio embeddings, ``{\normalfont ASR}'' means ASR embedding and confidence scores, while ``{\normalfont QE}'' includes QE score and its embedding. Input language feature is always used.}
  \label{tab:feature}
  \begin{tabular}{lcc}
    \toprule
     & WER & F1 \\
    Feature Groups & [\%] & [\%]\\
    \midrule
    Audio + ASR &12.4 & 61.5 \\
    Audio + QE &11.8 & 61.5\\
    ASR + QE &11.7 & 62.9\\
    Audio + ASR + QE (all) & \textbf{11.6} & \textbf{63.4} \\
    \midrule
    \enspace + QE rescoring     & 11.1 & 65.5 \\
    \bottomrule
  \end{tabular}
\end{table}

Table \ref{tab:feature} presents an ablation study about the impact of various features on predicting ASR performance, categorized into three groups: audio, ASR, and QE.
All cases with QE features exhibit a clear improvement in WER compared to the case without.
This underscores the value of QE scores and embeddings, which offer useful information distinct from audio or ASR features.
Comparing the second and third rows reveals a similar effect between audio and ASR features.

\begin{figure}[!h]
\centering
\includegraphics[width=\linewidth]{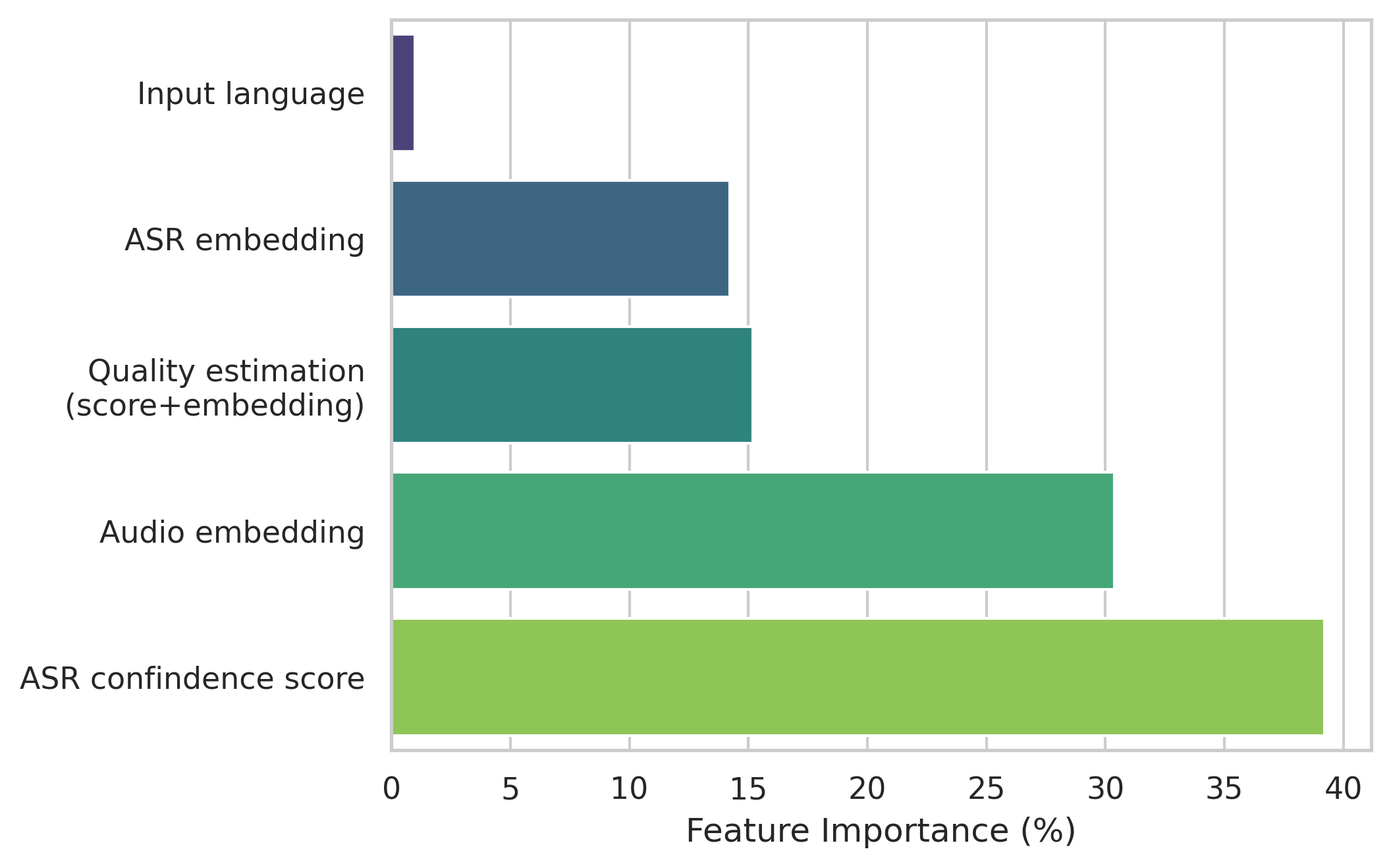}
\caption{Mean feature importance of binary classifiers.}
\label{fig:feature_importance}
\end{figure}

The optimal configuration undoubtedly involves combining all features. Figure \ref{fig:feature_importance} illustrates the feature importance computed within the GBM.
ASR confidence scores are deemed the most important, followed by embeddings from self-supervised audio models, quality estimation models, and ASR models.
This shows a considerable reliance on neural encoders for performance.
Interestingly, language categorization appears to hold minimal importance, highlighting the AutoMode-ASR versatility across languages.

\section{Conclusion}
This work introduces AutoMode-ASR, a novel framework designed to dynamically select the most suitable ASR system for a given audio input; which harnesses the strengths of different ASR technologies to substantially improve transcription accuracy.
It also considerably saves computational resources and operational costs by conducting binary system comparisons with a cost-effective system as the pivot.
Through rigorous testing, AutoMode-ASR shows remarkable adaptability across audio environments and linguistic contexts, reducing WER from 13.4\% to 11.1\% with 65\% lower cost and 75\% faster speed.
We verify that the multi-system ASR is a promising and practical way to optimize performance cost-effectively and time-efficiently through smart system selection.

%
%
\bibliographystyle{splncs04}
\bibliography{main}

\end{document}